\documentclass[a4paper,10pt]{article}
\usepackage{geometry}
\usepackage[utf8]{inputenc}
\usepackage{amsmath}
\usepackage{amssymb}
\usepackage{latexsym}
\usepackage{enumerate} 
\usepackage{amsthm}
\usepackage{qtree}

\usepackage[english]{babel}
\usepackage{setspace}
\usepackage{braket}
\usepackage{graphicx}
\usepackage{linguex}
\usepackage{bussproofs}
\usepackage{lscape}
\usepackage{tikz}
\usepackage{graphicx}
\usepackage{leftidx}
\usepackage[hyperref]{naaclhlt2018}
\usepackage{makecell}
\usepackage{natbib}
\usepackage{wrapfig}
\usepackage{changepage}
\usepackage{caption}

\aclfinalcopy 

\usepackage{footnote}
\makesavenoteenv{tabular}

\usepackage{lscape}
\usepackage{tikz}
\usepackage{graphicx}
\usepackage{leftidx}
\usepackage{tikz-cd}
\usepackage{listings}
\usepackage{mathtools}
\usepackage{blkarray, bigstrut}
\usepackage{tikz-dependency}
\usepackage{fancyhdr}

\tikzset{>=latex}
\usepackage{etoolbox}
\makeatletter
\patchcmd\@combinedblfloats{\box\@outputbox}{\unvbox\@outputbox}{}{\errmessage{\noexpand patch failed}}
\makeatother

\def\layersep{1.5cm}

\newcommand{\norm}[1]{\left\lVert #1 \right\rVert}

\usepackage[autostyle, english = american]{csquotes}
\MakeOuterQuote{"}

\begin{document}
\title{Augmenting Compositional Models for Knowledge Base Completion Using Gradient Representations}
\author{Matthias Lalisse\\
Dept of Cognitive Science\\
Johns Hopkins University\\
Baltimore, MD USA \\
{\tt lalisse@jhu.edu} \And
Paul Smolensky\\
Dept of Cognitive Science\\
Johns Hopkins University\\
\& Microsoft Research AI \\
Seattle, WA USA\\
{\tt smolensky@jhu.edu}
}

\maketitle

\begin{abstract}
Neural models of Knowledge Base data typically employ compositional representations of graph objects: entity and relation embeddings are systematically combined to evaluate the truth of a candidate knowledge base entry. Using a model inspired by Harmonic Grammar, we propose to tokenize triplet embeddings by subjecting them to a process of optimization with respect to learned well-formedness conditions on knowledge base entries. The resulting model, known as Gradient Graphs, leads to sizable improvements when implemented as a companion to compositional models. The "supracompositional" triplet token embeddings it produces have interpretable properties that prove helpful in performing inference on the resulting representations.\end{abstract}

\section{Introduction} 

As they are conventionally analyzed, representations of semantic or linguistic data are "compositional": the meanings of complex representations are built up from the meanings of their constituent parts. This idea has motivated numerous models of graph data deployed in knowledge base completion (KBC), in which embeddings of entities and relations are combined into composite representations|pairs of entities in a particular relation with one another|that are built up systematically from the constituent parts. But what happens when the whole is \emph{not} a simple function of the parts? A natural case arises in the interpretation of Noun-Noun compounds. The contrasting senses of \emph{vampire cat} (\emph{a-cat-that-is-a-vampire}) and \emph{vampire stake} (\emph{a-stake-used-to-kill-a-vampire}) has as much to do with the compatibility of the contituent nouns occurring in a given relation than with the meanings of the individual constituents.

Pursuing this line of thought, we propose \textbf{Gradient Graphs}, a neural network model for KBC built on the principle that compositionally-obtained representations of semantic objects can be optimized to reflect context-specific aspects of the meanings of their constituents. The issue of context-conditioned, tokenized semantic representations has received little explicit attention in the KBC literature. However, precedents do exist. \citet{bordes2011transe} model context-sensitive entity senses by embedding relations as pairs of matrices $(R_{lhs},R_{rhs})$ that linearly transform entity embeddings into pairs of embeddings defined by the relation and the entities' positions within it (the left-hand-side or right-hand-side). The distances of the resulting embeddings are then compared. \citet{socher2013neuraltensor} cope with the context-sensitivity of relation meanings by learning a $k\times d \times d$-dimensional tensor embeddings for each relation, letting their model represent polysemy by learning $k$ versions of the relation represented in the $k$ slices of its embedding tensor. The intuition underlying this approach is that, for instance, the relation \texttt{has\_{}part} has a different sense when applied to a biological organism than when predicated of a company. While the former has parts like organs and limbs, the latter has parts like subsidiaries and workers, which occupy very different parts of the semantic space. Each relational slice is then responsible for learning the compatibility of arguments within particular semantic subspaces.

In contrast to these other works, our approach is more radical in the sense that our context-sensitive representations of knowledge base entries are not just computed from the entries' constituent elements (entity and relation embeddings), but are instead the result of a representation-optimization procedure that balances compositionally-derived representations with general knowledge about the characteristics of well-formed semantic structures. We show that this additional "supracompositional" processing, in addition to yielding sizable accuracy improvements over the compositional models we apply it to, leads to embeddings of entity tokens with interpretable characteristics.

\subsection{Layout of the paper} 

Section \ref{sec_framework} lays out the general framework, which is compatible with a variety of implementations. Section \ref{sec_models} presents two compositional embedding models proposed in the literature. We adapt these models to construct compositional embeddings, and in Section \ref{sec_experiments} report evaluations of Gradient versions of these models. Section \ref{sec_discussion} discusses the characteristics of the resulting semantic representations in greater detail, as well as their role in assisting inference. Section \ref{sec_conclusion} concludes. Technical details about the model and the implementations are given in the Appendices.

\section{Optimization of semantic tokens} \label{sec_framework}

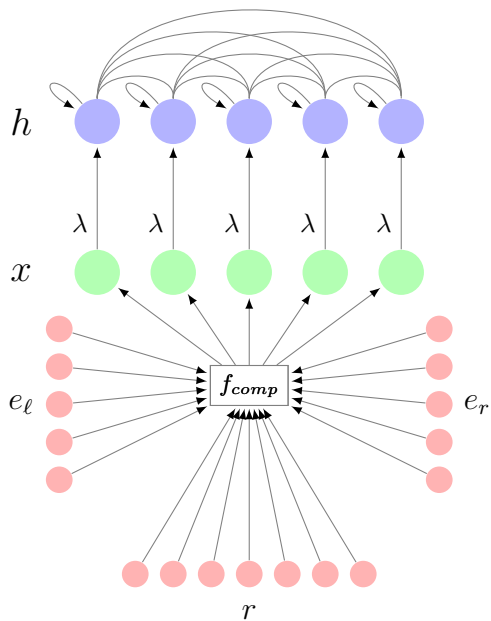
\begin{figure}[!h]
\centering
\begin{tikzpicture}[shorten >=1pt,draw=black!50, node distance=\layersep]
    \tikzstyle{neuron}=[circle,fill=black!25,minimum size=17pt,inner sep=0pt]
    \tikzstyle{tinyneuron}=[circle,fill=black!25,minimum size=10pt,inner sep=0pt]
    \tikzstyle{input neuron}=[neuron, fill=green!30];
    \tikzstyle{hidden neuron}=[neuron, fill=blue!30];
    \tikzstyle{embedding neuron}=[tinyneuron, fill=red!30];
    \tikzstyle{annot} = [text width=6em, text centered]
    \foreach \name / \x in {1,...,5}
        \node[input neuron,label] (I-\name) at (\x,0) {};

    \foreach \name / \x in {1,...,5}
        \path[yshift=0.5cm]
            node[hidden neuron] (H-\name) at (\x cm,\layersep) {};
            
    \foreach \source in {1,...,5}
        \path[->] (H-\source) edge [out=130,in=150,looseness=15] (H-\source);
    
    \foreach \source in {1,...,5}            \path (I-\source) edge[->] node[near start,left] {$\lambda$} (H-\source);
        \foreach \source in {1,...,5}
        \foreach \dest in {\source,...,5}
            \path [bend left=90] (H-\source) edge (H-\dest);
	\foreach \name / \x in {0,...,4}
        \node[embedding neuron] (E1-\name) at (.5, -2.75+\x*.5) {};
        
    \foreach \name / \x in {0,...,4}
        \node[embedding neuron] (E2-\name) at (5.5, -2.75+\x*.5) {};
     
    \foreach \name / \x in {-1,0,...,4,5}
        \node[embedding neuron] (R-\name) at (2+\x*.5,-4) {};
        
    \node[draw] (fcomp) at (3,-1.5) {\small $f_{comp}$};
	\foreach \source in {0,...,4}
            \path (E1-\source) edge[->] (fcomp); 
            
    \foreach \source in {0,...,4}
            \path (E2-\source) edge[->] (fcomp); 
            
    \foreach \source in {-1,...,4,5}
            \path (R-\source) edge[->] (fcomp); 
 
	\node[draw] (fcomp) at (3,-1.5) {\small $f_{comp}$};
	\foreach \dest in {1,...,5}
            \path (fcomp) edge[->] (I-\dest); 

    \node[align=left] at (0,0) {\Large $x$};
    \node[align=left] at (0,2) {\Large $h$};
    \node[align=right] at (0,-1.75) {\large $e_\ell$};
    \node[align=left] at (6,-1.75) {\large $e_r$};
    \node[align=left] at (3,-4.5) {\large $r$};
\end{tikzpicture}
\caption{Gradient Graph as a recurrent neural network. In addition to bias terms (omitted in the figure) and self-connections, hidden units are densely connected to one another via a layer of connections with symmetric (undirected) weights, and receive constant input weighted by $\lambda$ from a single unit in the input layer. The composition function $f_{comp}(e_\ell,r,e_r)$, which differs between implementations, computes a \textbf{compositional embedding} $x$, which is fed into a \textbf{hidden layer} $h$ of the network. The continuous-time dynamics of this network compute an internal representation $\hat{h}$ of the input triplet that is optimal with respect to the Harmony \ref{def_harmony}|a measure of the triplet's semantic well-formedness. } \label{fig_network_diagram}
\end{figure}

The hypothesis underlying the approach we propose is that noncompositional effects in knowledge base data can be modeled by subjecting candidate facts to a process of optimization with respect to a set of learned semantic coherence conditions. These semantic coherence conditions, encoded in a symmetric matrix, map out the covariance structure of the semantic space, indicating which semantic features are likely to co-occur with one another. The embedding of a given triplet is then the vector obtained by optimizing the semantic coherence of the the triplet embedding. 

We first lay out the model in abstract form, before introducing particular implementations. Let $x \in \mathbb{R}^d$ be a $d$-dimensional embedding of a knowledge base triplet $(e_\ell, r, e_r)$ obtained as some function $f_{comp}$|the \textbf{composition function}|of the embeddings of the left and right entities as well as the relation $r$. Section \ref{sec_models} provides several models for constructing the triplet embedding $x$. Also, let $h$ be a $d$-dimensional vector giving the internal ("hidden") state of the network. The \emph{Harmony} of an internal state $h$ of the network with respect to the triplet embedding $x$ is
\ex. {\small $\mathcal{H}(h,x) = \frac{1}{2} \left[ {\color{blue!60} h^\top \mathbb{W} h + b^\top h } {\color{red!70} - \lambda (h - x)^\top (h-x) } \right]$ } \label{def_harmony}

where $\mathbb{W}$ is a $d \times d$ weight matrix with $\mathbb{W}= \mathbb{W}^\top$ and $b$ is a bias vector, both learned. \ref{def_harmony} is composed of two terms: {\color{blue!60} \textbf{Core Harmony}}, a measure of the semantic coherence of the state vector $h$, and {\color{red!70} \textbf{Faithfulness}}, a penalty incurred due to the state $h$'s deviation from the compositional triplet embedding $x$. $\lambda$ is a hyperparameter that controls the magnitude of the penalty incurred for straying from $x$. 

$\mathcal{H}(h,x)$ may be rewritten as \ref{def_harmony_rewrite}. 
\ex. $\mathcal{H}(h,x) = \frac{1}{2} [ h^\top \left(\mathbb{W} - \lambda I \right) h + (b + 2\lambda x)^\top h $\\\phantom{h}\hfill$- \lambda x^\top x)]$ \label{def_harmony_rewrite}

If $\lambda$ is greater than the largest eigenvalue of $\mathbb{W}$, then $V = \mathbb{W}- \lambda I$ is negative-definite, and $\mathcal{H}(h,x)$ has a unique global optimum $\hat{h} = \text{argmax}_h \mathcal{H}(h,x)$ for each $x$. In closed form, this global optimum is
\ex. $\mu(x) =  -V^{-1}\left(\frac{1}{2} b + \lambda x\right) $ \label{def_mu_x}

which depends only on the network parameters and on $x$. The expression $\mu(x)$ comes from observing that $\hat{h}$ is the mean of a Gaussian distribution with inverse covariance matrix $V$, which implies that $\hat{h}$ is the most probable state $h$ of the network with respect to the probability distribution over the state space defined by $p(h|x) \propto \exp\{\mathcal{H}(h,x)\}$ (see Appendix A). We take the token embedding for a triplet $x$ to be $\mu(x)$, which is the most semantically coherent triplet embedding given the compositional triplet $x$. In the limit as $\lambda \rightarrow \infty$, $\mu(x)$ is just $x$ itself. Let $\lambda_{\mathbb{W}}$ denote the largest eigenvalue of $\mathbb{W}$; then as $\lambda \rightarrow \lambda_{\mathbb{W}}$, $\mu(x)$ may become arbitrarily far from the triplet embedding $x$.

A Gradient Graph may be viewed as a neural network with weight matrix $\mathbb{W}$ and bias vector $\frac{b}{2}$, where the synaptic weights $\mathbb{W}$ specify a feedback layer through which the values of the hidden state units affect one another. The construction is as follows. We stipulate that the hidden state of the network follows the gradient of Harmony over time:
\ex. $\frac{d h}{dt} = \frac{\partial \mathcal{H}(x,h)}{ \partial h }$

Therefore, 
\begin{align*}
\frac{d h_i}{dt} &= \frac{d}{dh_i} \frac{1}{2} \left[h^\top \mathbb{W} h + b^\top h - \lambda \norm{x - h}^2 \right] \\
	&= \frac{d}{dh_i} \frac{1}{2} \left[ \sum_{jk} h_j\mathbb{W}_{jk}h_k + b_ih_i - \lambda (x_i - h_i)^2 \right] \\
	&= \frac{1}{2} \left[ \sum_j h_j \mathbb{W}_{ji} + \sum_k \mathbb{W}_{ik} h_k \right] + \frac{b_i}{2} + \lambda x_i - \lambda h_i 
\end{align*}
The above specifies the connectivity of a network whose hidden units have the linear transfer function ($f(input) = input$), bias $\frac{b}{2}$ and external input $x$ (weighted by $\lambda$). Each $h_i$ also receives self-inhibitory input weighted by $-\lambda$, as well as inputs $\mathbb{W}_{ij} h_j$ from each $h_j$. The symmetry of $\mathbb{W}$ implies that each term $\mathbb{W}_{ij}h_j = h_j\mathbb{W}_{ji}$ occurs twice, so that the factor of $\frac{1}{2}$ cancels. This connectivity structure is illustrated in Figure \ref{fig_network_diagram}. 

\subsection{Relation to Harmonic Grammar}
In addition to being globally optimal with respect to the Harmony function $\mathcal{H}(h,x)$ conditioned on a particular input $x$, $\mu(x)$ is the unique fixed point of this network's state-evolution dynamics. It is interesting to note that such networks are the connectionist foundation for Harmonic Grammar (HG) and Optimality Theory (OT) in Linguistics \cite{smolensky_legendre2006harmonic_mind}, where the dynamics of a neural network perform optimization over internal representations of an input structure. Appropriate output representations are then selected in accordance with well-formedness constraints encoded in the network parameters. There, the output representation balances Faithfulness to the input (an Underlying Form) and the network's knowledge about the characteristics of well-formed structures in general. 

Similarly, it is appealing to conceptualize the hidden layer of a Gradient Graph network as cleaning up a knowledge base triplet by subjecting it to semantic well-formedness conditions. The optimal triplet $\mu(x)$ is then the point to which the network converges in the limit of infinite computation time. However, our model differs from typical implementations of HG and OT in that the optimal structure $\mu(x)$ does not, in general, decompose into a unique combination of the input constituents (entity and relation embeddings). The resulting representations are in this sense gradient, rather than being the product of a combination of discrete objects. Furthermore, Gradient Graphs are, to our knowledge, the first application of these ideas to the automatic learning of an appropriate semantic optimization function from a large amount of data.

\subsection{Comparison with translation-based approaches}

Like a large class of \emph{Translation}-based models \citep{bordes2011transe,yoon2016translation,lin2015transR,ji2016sparseTrans}, our inference procedure consists of the application of an affine transformation to an input $x$ (Equation \ref{def_mu_x}), which is then scored using some regular operation. In our case, this scoring function is quadratic. A particular close cousin is the bilinear \textsc{Semantic Matching Enery} (\textsc{SME}) method of \citet{bordes2014semanticenergy}, which learns a global third-order tensor $\mathbb{W}$ that, when dotted along the third mode with a relation embedding $r$, yields a relation-specific matrix $\mathbb{W}_r$. Along with learned left and right bias vectors $b_\ell$ snd $b_r$, this weight matrix is fed into the bilinear scoring function \ref{def_score_SME}:
\ex. $\text{score}_\textsc{SME}(e_\ell, r, e_r) = $\\
\phantom{h}\hfill $(\mathbb{W}_r e_\ell + b_\ell)^\top (\mathbb{W}_r e_r + b_r)$\label{def_score_SME}

Expanding out this expression, we get \ref{sme_expanded}.
\ex. ${e_\ell}^\top {\mathbb{W}_r}^\top\mathbb{W}_r e_r + {b_\ell}^\top \mathbb{W}_r e_r + {b_r}^\top \mathbb{W}_r e_l + {b_\ell}^\top b_r $
\label{sme_expanded}

The relation-specific bilinear form ${\mathbb{W}_r}^\top \mathbb{W}_r$ is, like our global $\mathbb{W}$ matrix, symmetric. The remaining terms, apart from the constant ${b_\ell}^\top b_r$, compute a pair of relation-specific bias vectors ${b_\ell}^\top \mathbb{W}_r$ and ${b_r}^\top \mathbb{W}_r$ applied to the pair of entity embeddings.	The resulting Energy function used to score triplets has more than a passing similarity to our Harmony function \ref{def_harmony} when $\lambda = \infty$ and, thus, no optimization takes place. 

A distinctive characteristic of our approach in relation to these structurally similar models is that the transformation undergone by a Gradient Graph triplet is directly connected to the well-formedness criterion according to which triplets are evaluated in inference. As illustrated in the Discussion, our transformation of a compositional triplets using learned well-formedness criteria leads to two kinds of triplet embeddings: compositionally obtained \emph{type} embeddings, and contextually optimized \emph{token} embeddings. In qualitative and quantitative analyses of the learned representation, we see (1) that the space of compositionally obtained triplet embeddings has a reasonable structure, independently of the optimizing transformation, that is already sensitive to the context supplied by the relation, and (2) that semantic optimization improves these compositional representations in recognizable ways. Interestingly, improving triplets with respect to the Harmony function does \emph{not} uniformly place them in regions that are high-Harmony in a global sense. In fact, we find that whereas positive triplets end up close to other positive triplets, plausible but negative triplets tend to be detained in clusters with other negative instances (Tables \ref{table_neighborhood_results} and \ref{table_sem_neighborhoods}). 

\section{Compositional and Gradient Models} \label{sec_models}

Optimization with respect to $\mathcal{H}$ can be implemented wherever we can construct a triplet embedding $x$. In our experiments, we apply Harmonic optimization of triplet representations to two compositional embedding models drawn from the knowledge base completion literature: \textsc{DistMult} and \textsc{HolE}. Both models specify a scoring function for triplet embeddings obtained via operations applied to embeddings of the three triplet components|two entity vectors and a relation vector|with no additional learned components apart from these representations of the triplet constituents. We take the terms occurring in these scoring functions to be components of the \emph{representation} of the triplet, specifying what information about the triplet elements is important to evaluating the triplet's quality. Hence, we constructed Harmonic triplet embeddings according to the desideratum that every term occurring in the basic method's scoring function should also appear in the triplet representation $x$ in the Harmonic model. 
For instance, the score of a \textsc{DistMult} triplet is a sum of three-way products of the corresponding elements of the embeddings $e_{\ell}$, $r$, and $e_r$. Setting the products $[e_\ell]_i [r]_i [e_r]_i$ to appear in our compositional triplet embeddings (as in Eqn \ref{def_x_hdistmult}) satisfies this desideratum. 

\textsc{DistMult} \cite{yang2015distmult} is a baseline model for scoring knowledge base triplets using the scoring function \ref{def_score_distmult}:
\ex. $score_{\textsc{DistMult}}(e_\ell,r,e_r) = {e_\ell}^\top diag(r) {e_r}$ \label{def_score_distmult}

where $e_\ell, r, e_r$ are $d$-dimensional embeddings and $diag(r)$ is the $d\times d$-dimensional matrix obtained by arranging the elements of $r$ along the diagonal. \citet{kaldec2017baselines} have recently shown that \textsc{DistMult} can outperform many more complicated scoring functions when hyperparameters are properly optimized, making it a strong baseline comparison for the method we propose. In addition, \textsc{DistMult} often occurs as a subcomponent in state-of-the-art KBC models|e.g. \citep{schlichtkrull2017graphconv,toutanova2015jointembedding}. From this starting-point, we construct \textsc{Harmonic DistMult} (\textsc{HDistMult}) by setting the triplet embedding $x$ to the elementwise multiplication of the relation and the pair of entity vectors:
\ex. $x_{\textsc{HDM}} = e_\ell \odot r \odot e_r$ \label{def_x_hdistmult}

where $\odot$ denotes elementwise multiplication. 

\textsc{Holographic Embeddings} (\textsc{HolE}) were introduced by \citet{nickel2016hole} building on theoretical work by \cite{plate1995holographic}, as a means of constructing compressed tensor product representations of relational triplets. The method computes the score for a triplet $(e_\ell, r, e_r)$ from the similarity between a relation vector and the circular correlation $e_\ell \star e_r$ of the entity vectors and a relation vector: 
\ex. $score_{\textsc{HolE}}(e_\ell, r, e_r) = r^\top (e_\ell \star e_r)$ \label{def_score_HolE}

where the circular correlation of $e_\ell$ and $e_r$ is computed as \ref{def_circular_correlation}. 
\ex. $e_\ell \star e_r = \mathcal{F}^{-1}\left( \overline{\mathcal{F}(e_\ell)} \odot \mathcal{F}(e_r)\right)$ \label{def_circular_correlation}

$\mathcal{F}$ and $\mathcal{F}^{-1}$ denote the Fourier Transform and its inverse, and $\overline{\mathcal{F}(e_\ell)}$ is the complex conjugate of $\mathcal{F}(e_\ell)$.\footnote{
The Fourier transform decomposes a function of time into its frequency components. In the context of holographic embeddings, its utility comes from the \emph{Convolution Theorem}, which states that convolution in the time domain corresponds to elementwise multiplication in the frequency domain. This is useful in actual computations. The circular correlation|which consists of convolution with a time-reversed signal|can also be computed as a sum over off-diagonals of the tensor product of vectors, with time complexity $\mathcal{O}(d^2)$. In contrast, the Fast Fourier Transform (FFT) has time complexity $\mathcal{O}(n\log n)$ \citep{nickel2016hole}.  \label{note_fourier}
}
 Circular correlation is asymmetric ($e_\ell \star e_r \not = e_r \star e_\ell$)|allowing it to model asymmetric relations|and the result of the operation has the same dimensionality as the input vectors, while still carrying information about which pair of entities was bound together via correlation. 

We construct \textsc{Harmonic HolE} (\textsc{HHolE}) triplet embeddings via elementwise multiplication of relation vectors with the correlated pair of entity vectors:
\ex. $x_{\textsc{HHolE}} = r \odot (e_\ell \star e_r)$ \label{def_x_HHolE}

In both Harmonic models, the score for a candidate triplet $(e_\ell, r, e_r)$ with embedding $x$ is calculated by taking the Harmony of its optimal instantiation, $\hat{h} = \mu(x)$, i.e.
\ex. $score(x) = \mathcal{H}(\mu(x),x)$

In the experiments, we train our networks using the log-softmax objective with negative sampling. For each positive training example $(e_\ell, r, e_r)$ with embedding $x$, we construct $N$ negative examples $(\tilde{e}^n_{\ell}, \tilde{r}^n, \tilde{e}^n_r)$ obtained by deleting either the left or right entity of the true triplet and replacing it with a randomly sampled entity vector. Let $\tilde{x}^n$ denote the embedding of the $n$th negatively sample triplet $(\tilde{e}^n_{\ell}, \tilde{r}^n, \tilde{e}^n_r)$. The training objective is then to minimize \ref{def_harmony_loss}:
\ex. {\small $\mathcal{L}_{\mathcal{H}}(e_\ell, r, e_r) = $\\
	\phantom{h}\hfill $-\log \frac{\exp \{ \mathcal{H}(\mu(x),x)\}}{\exp \{ \mathcal{H}(\mu(x),x)\} + \sum_{n=1}^N \exp \{ \mathcal{H}(\mu(\tilde{x}^n), \tilde{x}^n) \}}$ }\label{def_harmony_loss}

This has the effect of increasing the Harmony of positive examples relative to negative samples. The learning rule is thus Harmony-maximizing: the network parameters maximize the well-formedness of the positive examples relative to negative samples.

\setlength{\extrarowheight}{2pt}
\noindent{
\begin{table*}[h!]
\begin{adjustwidth}{-.25cm}{}
\captionsetup{width=.95\textwidth}
\centering
{\scriptsize 
\begin{tabular}{@{\extracolsep{5pt}}lcccccccccccc@{}}
  & \multicolumn{6}{c}{\textbf{FB15K}} & \multicolumn{6}{c}{\textbf{WN18}} \\ \cline{2-7} \cline{8-13}
& & \multicolumn{2}{c}{\textbf{Rank}} & \multicolumn{3}{c}{ \textbf{Hits@}} & & \multicolumn{2}{c}{\textbf{Rank}} & \multicolumn{3}{c}{\textbf{Hits@}} \\ \cline{3-4} \cline{5-7} \cline{9-10} \cline{11-13} 
{\textbf{Model} }&$\mathbf{\lambda}$ & \textbf{MR} & \textbf{MRR} & \textbf{1} & \textbf{3} & \textbf{10}  
& $\lambda$ & \textbf{MR} & \textbf{MRR} & \textbf{1} & \textbf{3} & \textbf{10} \\ \cline{1-1} \cline{2-2}\cline{3-3}\cline{4-4}\cline{5-5}\cline{6-6}\cline{7-7}\cline{8-8}\cline{9-9}\cline{10-10}\cline{11-11}\cline{12-12}\cline{13-13}
\textsc{DistMult} & - & - & .350 & - & - & .577 
& - & - & .830 & - & - & .942\\ 
\textsc{Ensemble DM}$^\dagger$ & - & 36 & \textbf{\underline{.837}} & \textbf{\underline{.797}} & - & \textbf{\underline{.904}}	& 	
 - & 457 & .790 & \textbf{.784} & - & .950  \\
\textsc{DistMult}$^*$ & - & 28 & .710 & .605 & .792 & .876 
& - & 220 & .825 & .714 & .938 & .950 \\ 
\textsc{HDistMult} & $\infty$ & \textbf{23} & .806 & .751 & \textbf{.845} & .898 
&$\infty$ & \underline{\textbf{164}} & \textbf{.841} & .740 & \textbf{.943} & \underline{\textbf{.955}} \\ 
\textsc{HDistMult} &$50.0$ & \textbf{23} & .742 & .661 & .799 & .881
 & $3.0$ & 184 & .831 & .732 & .931 & .945 \\ \hline
\textsc{HolE} & - & - & .524 & .402 & .613 & .739 
& - & - & .938 & .930 & \underline{\textbf{.945}} & .949 \\  
\textsc{HolE}$^*$ & - & 39 & .409 & .289 & .464 & .647
& - & 205 & .916 & .893 & .936 & .946 \\ 
\textsc{HHolE} & $\infty$ & 32 & .682 & .575 & .763 & .850  
&$\infty$ & 293 & .919 & .903 & .934 & .942 \\ 
\textsc{HHolE} & $1.0$ & \underline{\textbf{21}} & \textbf{.796} & \textbf{.727} & \underline{\textbf{.848}} & \underline{\textbf{.901}}  
&$2.0$ & \textbf{183} & \underline{\textbf{.939}} & \underline{\textbf{.931}} & \underline{\textbf{.945}} & \textbf{.951} \\
\end{tabular}
}
\caption{Results on FB15K and WN18. The results from the original \textsc{DistMult} and \textsc{HolE} models are drawn from \citep{yang2015distmult} and \citep{nickel2016hole}. Our reimplementations$^*$ of \textsc{DistMult} and \textsc{HolE} differ in numerous details from those in the original papers (see Appendix B for technical details). \textsc{Ensemble DistMult}$^\dagger$ refers to the hyperparameter-optimized Ensemble (product of experts) reimplementation of DistMult proposed by \citet{kaldec2017baselines}. For each model, we report Mean Rank (MR) and Mean Reciprocal Rank (MRR), as well as Hits@$N$ for $N\in \{1,3,10\}$. Hits@$N$ denotes the fraction of test instances in which the true triplet completion had rank less than or equal to $N$. The best results within each category (\textsc{DistMult} and \textsc{HolE}) are marked in \textbf{bold}, and the best results overall are additionally \underline{\textbf{underlined}}. } \label{table_results}
\end{adjustwidth}
\end{table*} 
}

\section{Experiments} \label{sec_experiments}

We evaluated Gradient Graphs using the standard WN18 and FB15K datasets \citep{bordes2013multirelational}|which are subsets of the WordNet \citep{wordnet1995} and Freebase \cite{freebase2008} databases|on the Entity Reconstruction task. In Entity Reconstruction, the network ranks completions of triplets $(\ \cdot\ , r, e_r)$ and $(e_\ell, r, \ \cdot\ )$ with deleted left and right entities. The model is successful if it ranks the true triplet above other candidate completions. We report results in the \emph{filtered} evaluation setting \citep{bordes2013multirelational}, in which a test triplet is only ranked against triplets that do not occur in the database. The rank of a test triplet is thus the rank of the \emph{first} correct answer to the query. For both \textsc{DistMult} and \textsc{HolE}, we report the  originally reported results alongside results for our reimplementations, comparing these models with our Harmonic variants \textsc{HDistMult} and \textsc{HHolE} with and without optimization of hidden layer representations. The Harmonic models with $\lambda = \infty$ have the Harmony function $\mathcal{H}(x,x)$, i.e. where the hidden representation is just the compositional embedding itself and the Faithfulness penalty in \ref{def_harmony} is 0. 

Our models used 256- to 512-dimensional embeddings and manually tuned values of the hyperparameter $\lambda$. In all models, entity and relation embeddings were normalized to $\norm{v} = 1$. We do not regularize parameters, but instead set an upper bound $\lambda - \epsilon$ ($\epsilon$ a small constant) on the $l_2$ norm of the weight matrix $\mathbb{W}$, which helps constrain the spectral norm (maximum eigenvalue) of $\mathbb{W}$ to remain lower than $\lambda$. This may be seen as adopting a uniform prior on weight matrices lying within the $n$-ball with squared radius $\lambda - \epsilon$. Importantly, this procedure keeps the matrix $V = \mathbb{W} - \lambda I$ negative-definite|a necessary condition for the existence of a unique optimum for $\mathcal{H}(h,x)$. 

Results from the experiments are reported in Table \ref{table_results}. 
Overall, we found that models using our quadratic scoring function \ref{def_harmony} to perform best across the board. This effect was particularly seen in more stringent evaluation criteria|Hits@1 and Hits@3|leading to, for instance|a 15\% improvement in Hits@1 (accuracy) on Freebase between our \textsc{DistMult} reimplementation and quadratic \textsc{HDistMult} ($\lambda = \infty$). The best\textsc{DistMult} models were those with high $\lambda$ values; however, within-model comparison of \textsc{HolE} shows dramatic improvements from including the optimization component|a 32\% increase in FB15K accuracy between the results of \citep{nickel2016hole} and our \textsc{HHolE} with a permissive $\lambda$-criterion of 1.0. 

\section{Discussion} \label{sec_discussion}

In part, the appeal of our supracompositional representations stems from their ability to produce emeddings of \emph{tokens} of semantic objects|that is, embeddings that take into account the context of a particular instance of a semantic type. Tokenized embeddings have proven useful in various settings. For instance, \citet{dasigi2017tokenembeddings} construct token embeddings by superposing learned vectors for WordNet senses in ratios determined by a probability distribution computed from the context. The resulting representation is a context-weighted sum of discrete senses drawn from a hand-crafted ontology. Closer to our approach, \citet{belanger2015lds} model text as a linear dynamical system that generates texts through transitions of a continuous-state, discrete-time dynamical system across time. Estimates of the system's most probable internal state can then be extracted as an embedding of the tokens, which prove useful in language modeling and other downstream tasks.

In our framework, \emph{types} correspond to static entity and relation embeddings that are the input to $f_{comp}$, and the triplet embeddings resulting from their combination. \emph{Token} triplet embeddings are produced by optimization of the hidden layer of a \textsc{GG}. 
To understand the effect of optimizing the hidden layer of a \textsc{GGraph} both on its learned representations and on its performance in inference, we used the best-performing trained \textsc{HHolE} model to produce token embeddings of database triplets in order to inspect their semantic neighborhoods. 

For a given compositional triplet embedding $f_{comp}(e_\ell, r, e_r)\equiv x$, we first computed the optimized triplet representation $\mu(x) \equiv \hat{h} $ using Equation \ref{def_mu_x}. Treating $\hat{h}$ as the contextually optimal (token) embedding of the triplet $(e_\ell, r, e_r)$, we then examined the semantic neighborhood by computing the 5 closest optimized embeddings in the context of the same relation. Table \ref{table_sem_neighborhoods} shows the semantic neighborhoods of compositional triplets $x$ and optimized triplets $\hat{h}$ for different possible completions of a number of queries. Rows 1 and 2 display completions of the query $( \ \cdot\ , \mathtt{office\_{}position\_{}or\_{}title}, \mathtt{US\_{}President})$, and Rows 2 and 4 consider the neighborhood of the entity embedding of \emph{Bob Dylan} in the context of queries about his profession ($e_\ell = \mathtt{Bob\_{}Dylan}, r  = \mathtt{has\_{}profession}$) while varying the profession $e_r$. This illustrates how the representation of \emph{Bob Dylan} varies across his different professional guises. 

The table illustrates the utility of token embeddings in inference. Token embeddings of \emph{George W. Bush} and \emph{Barack Obama} in the context of a query about their having held the office of U.S. president are in semantic neighborhoods with a greater density of true instances of U.S. Presidents than their type embeddings. The negative examples \emph{John McCain} and \emph{Hilary Rodham Clinton} have type embeddings that are close to actual presidents. This is sensible since, for instance, Hilary Rodham Clinton is married to Bill Clinton|one of her nearest neighbors. But both of the negative examples' token embeddings have neighborhoods that are mostly cleared of actual presidents|despite having type embedding neighborhoods that are relatively dense with presidents. 

Turning to the second half of the table, we note that \emph{Bob Dylan}'s type embedding is already in a neighborhood dense with singer-songwriters. It is appropriate, then, that this neighborhood undergoes no change apart from minor re-ranking when the triplet (\texttt{Bob\_{}Dylan},\texttt{has\_{}profession},\texttt{singer-songwriter}) is optimized. For more difficult cases, however, where \emph{Dylan} is not a prototypical example, the semantic neighborhoods undergo dramatic reconfiguration. For instance, optimizing the triplet (\texttt{Bob\_{}Dylan},\texttt{has\_{}profession},\texttt{disc\_{}jockey}) correctly places \emph{Dylan} in the neighborhood of other DJs, despite the implausibility of this association in the neighborhood of his type embedding, which contains no DJs. This places him in the token neighborhood of \emph{Moby}, who is otherwise quite unlike \emph{Bob Dylan} except in respect of their common career as DJs. 

Combined with our finding that optimization yields the most dramatic improvements in the more stringent evaluation criteria (Hits@1 and Hits@3), this suggests that our optimization procedure is particularly helpful in arbitrating between difficult cases. This qualitative observation about the neighborhoods of compositional and supracompositional triplets can be quantified. Using the triplet classification dataset introduced by \citet{socher2013neuraltensor}, which contains an equal number of positive and negative triplets, we find (Table \ref{table_neighborhood_results}) that positive triplets, on average, end up in supracompositional neighborhoods that are more dense in positive examples than their compositional counterparts. On the other hands, negative triplets suffer a decrease in the number of positive triplets in the neighborhoods of their supracompositional embeddings. 
\begin{table}
\begin{center}
\begin{tabular}{|l|l|l|l|} \hline
	&	$\Delta$density & $t$ statistic & $p$ \\\hline
\textsc{Pos} 	& 0.241	& $t=99.7$ & $p \ll 10^{-10}$\\
\textsc{Neg}	& -0.059 & $t=-62.4$ & $p \ll 10^{-10}$ \\\hline
\end{tabular} \end{center}
\caption{Change in neighborhood (top-5 closest neighbors) density of true triplets ($\Delta\text{density}$) for positive and negative triplets drawn from the triplet classification dataset introduced by \citet{socher2013neuraltensor}, which is derived from the FB15K test set and consists of 59,071 positive triplets and the same number of negative triplets. This resulted in $N=118,142$ queries for both positive and negative examples (two for each triplet, querying both the left and right entity). After computing each triplet's neighborhood, we counted the number of triplet neighbors that were in fact in the training, validation, or test sets of FB15K, yielding a measure of the concentration of true and false examples in the neighborhood of both type and token triplet embeddings. } \label{table_neighborhood_results}
\end{table} 

To further quantify the role of semantic optimization in inference, we correlated the difference between the Harmony (score) of input triplets pre- and post-optimization with the change in its rank on the FB15K dataset. The change in Harmony is computed as $\Delta \mathcal{H} = \mathcal{H}(\mu(x),x) - \mathcal{H}(x,x)$, i.e. the difference between the Harmony of the token embedding and the Harmony of the type embedding. This comparison is model-internal|it does not compare models trained to do token inference with models trained for type inference. However, it serves as a useful index of the performance gains attributable to the optimization procedure. If optimizing a triplet representation indeed improves its relative position among all candidate triplets, we expect changes in Harmony to be negatively correlated with the change in rank of positive triplets. Consistent with this, we find that optimization leads to significant improvements in raw rank in our best trained \textsc{HHolE} model (Spearman's $\rho = -0.0157, p<10^{-6}$, Figure \ref{fig_optimization_effect}). When considering the change in Mean Reciprocal Rank, a more standard evaluation metric, we find that $\Delta\mathcal{H}$ is positively associated with improvements in MRR $\left(\rho = .1370, p\ll 10^{-10}\right)$,\footnote{$\Delta$MRR is computed as $\text{MRR}(\mu(x)) -\text{MRR}(x)$, i.e. the difference between the Mean Reciprocal Rank of the supracompositional and the compositional triplets.} particularly when triplets whose ranks do not change at all are omitted $\left(\rho = .3746, p\ll 10^{-10}\right)$. In other words, when semantic optimization makes a difference, it does so for the better. 

For \textsc{HDistMult}, $\Delta\mathcal{H}$ is significantly associated with increases in the rank of true triplets $\left( \rho = 0.1226, p\ll 10^{-10}\right)$, a result consistent with our finding that this class of models disprefers low settings of $\lambda$. This illustrates the importance of choices of representational format for embeddings of semantic data. Our optimization procedure can only operate over information that is contained in its compositional input. Hence, choices about how to combine the learned features of entities and relations|i.e. about the manner of composition|are central to our framework.
\begin{figure}
\includegraphics[scale=.45]{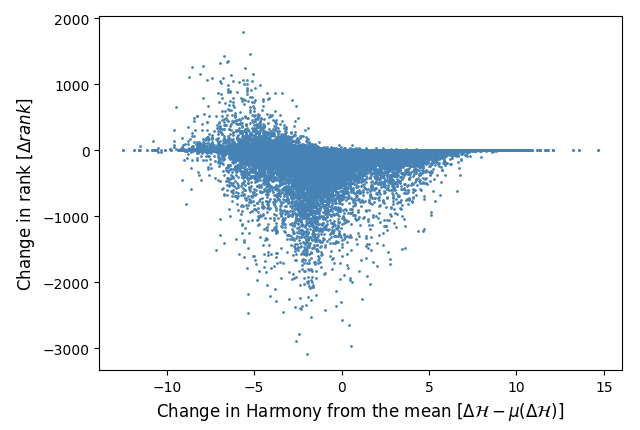}
\caption{Effect of optimization on the rank of FB15K validation set triplets ($N=100,000$; $50,000$ triplets with two queries per triplet) from the best-performing \textsc{HHolE} model ($d=512,\lambda=1.0$). The horizontal axis is a triplet's change in Harmony pre- and post-optimiation ($\Delta\mathcal{H} \equiv\mathcal{H}(\mu(x),x) - \mathcal{H}(x,x)$) minus the mean change in Harmony for all triplets ($\mu(\Delta\mathcal{H})$). This is plotted against $rank_q(\mu(x)) - rank_q(x)$, the triplet's change in rank due to optimization for query $q$. A negative correlation indicates reductions in rank (improvements) associated with increasing optimization of triplet representations.
}
\end{figure}\label{fig_optimization_effect}

\subsection{Desiderata of a composition function}
What factors affect the success of semantic optimization in combination with a particular composition scheme? We suspect that multiplicative interactions across embedding components|which are present in \textsc{HHolE} and absent in \textsc{HDistMult}|are essential for our optimization procedure to contribute helpfully to inference. Both \textsc{DistMult} and \textsc{HolE} are special cases of contracted Tensor Product Representations (\textsc{TPR}s), obtained by summing over (\textsc{HolE}) or discarding (\textsc{DistMult}) terms from the three-way tensor product $e_\ell \otimes r \otimes e_r$.\footnote{See \citep{nickel2016hole} for discussion of holographic embeddings as compressed tensor products.} In particular, \textsc{DistMult} retains only multiplicative interactions \emph{within} components, omitting terms with non-matching indices. This fact appears to be crucial. In a follow-up experiment, we implemented a series of full \textsc{TPR} models trained on FB15K, using the composition operation \ref{def_x_htpr}:
\ex. $x_{\textsc{HTPR}} = e_\ell \otimes r \otimes e_r$ 	\label{def_x_htpr}

Such models are necessarily small in size due to the rapid growth of dimensionality for \textsc{TPR}s as a function of the dimensionality of entity and relation embeddings. Consequently, their performance is also poor in comparison to our other implementations. However, the trend matched that which we observed within the \textsc{HHolE} class: models including the optimization procedure consistently outperformed those with $\lambda=\infty$ (see Table \ref{table_HTPR_results}). From this, we conclude that other embedding-based KBC models incorporating cross-component multiplicative interactions are likely to see improvements from the addition of a semantic optimization step prior to scoring.
\begin{table}
\begin{tabular}{l||l|l|l|l|l}
$\mathbf{\lambda}$ & \textbf{MR} & \textbf{MRR} & \textbf{H@1} & \textbf{H@3} & \textbf{H@10} \\\hline \hline
$\infty$ & 150 & .278 & .192 & .305 & .447	\\
$1.0$ & 134 & .295 & .204 & .326 & .471
\end{tabular}
\caption{Performance of \textsc{HTPR} models with and without optimization (controlled by $\lambda$). For both models, entities were $5$-dimensional and relations $20$-dimensional. This trend held across other hyperparameter settings.} \label{table_HTPR_results}
\end{table}

\begin{table*} 
\begin{tabular}{lll | lll} 
\multicolumn{6}{c}{ \emph{US Presidents} } \\ \hline\hline
 \multicolumn{3}{c}{\small \textbf{George W. Bush}} & \multicolumn{3}{c}{\small \textbf{Barack Obama}}  \\  \hline
 \multicolumn{1}{l}{\small $n$} & \multicolumn{1}{l}{\small \textbf{$x$} (compositional)} & \multicolumn{1}{l|}{\small \textbf{$\hat{h}$} (optimized)} & \multicolumn{1}{l}{\small $n$} & \multicolumn{1}{l}{\small \textbf{$x$} (compositional)} & \multicolumn{1}{l}{\small \textbf{$\hat{h}$} (optimized)}\\ \hline 
  {\scriptsize 1 } & { \scriptsize \textbf{George H. W. Bush} } & { \scriptsize \textbf{George H. W. Bush}} &  {\scriptsize 1 } & { \scriptsize Hillary Rodham Clinton } & { \scriptsize \textbf{George W. Bush}} \\ 
 {\scriptsize 2 } & { \scriptsize \textbf{Bill Clinton} } & { \scriptsize \textbf{Bill Clinton}} &  {\scriptsize 2 } & { \scriptsize Al Gore } & { \scriptsize \textbf{Bill Clinton}}\\ 
 {\scriptsize 3 } & { \scriptsize \textbf{Jimmy Carter} } & { \scriptsize \textbf{Jimmy Carter}} &  {\scriptsize 3 } & { \scriptsize \textbf{George W. Bush} } & { \scriptsize \textbf{John F. Kennedy}}\\ 
 {\scriptsize 4 } & { \scriptsize \textbf{John F. Kennedy} } & { \scriptsize \textbf{Ronald Reagan}} &  {\scriptsize 4 } & { \scriptsize \textbf{Bill Clinton} } & { \scriptsize \textbf{Ronald Reagan}}\\ 
 {\scriptsize 5 } & { \scriptsize \textbf{Ronald Reagan} } & { \scriptsize \textbf{Barack Obama}} &  {\scriptsize 5 } & { \scriptsize \textbf{John F. Kennedy} } & { \scriptsize \textbf{George H. W. Bush}}\\ \hline
 \\
 \multicolumn{3}{c}{\small John McCain} & \multicolumn{3}{c}{\small Al Gore}  \\ \hline
 \multicolumn{1}{l}{\small $n$} & \multicolumn{1}{l}{\small \textbf{$x$} (compositional)} & \multicolumn{1}{l|}{\small \textbf{$\hat{h}$} (optimized)} & \multicolumn{1}{l}{\small $n$} & \multicolumn{1}{l}{\small \textbf{$x$} (compositional)} & \multicolumn{1}{l}{\small \textbf{$\hat{h}$} (optimized)} \\ \hline 
 {\scriptsize 1 } & { \scriptsize John Kerry } & { \scriptsize John Kerry} &  {\scriptsize 1 } & { \scriptsize \textbf{Barack Obama} } & { \scriptsize Condoleezza Rice} \\
 {\scriptsize 2 } & { \scriptsize Hillary Rodham Clinton } & { \scriptsize Colin Powell}&  {\scriptsize 2 } & { \scriptsize \textbf{George W. Bush} } & { \scriptsize John C. Calhoun}\\ 
 {\scriptsize 3 } & { \scriptsize Colin Powell } & { \scriptsize Nancy Pelosi} &  {\scriptsize 3 } & { \scriptsize Colin Powell } & { \scriptsize Colin Powell }\\ 
 {\scriptsize 4 } & { \scriptsize \textbf{Richard Nixon} } & { \scriptsize Joe Biden} &  {\scriptsize 4 } & { \scriptsize Condoleezza Rice } & { \scriptsize Hillary Rodham Clinton}\\ 
 {\scriptsize 5 } & { \scriptsize \textbf{Herbert Hoover} } & { \scriptsize Dick Cheney } &  {\scriptsize 5 } & { \scriptsize \textbf{John F. Kennedy} } & { \scriptsize John Kerry }\\ \hline  
 \\
\multicolumn{6}{c}{ \emph{Guises of Bob Dylan } } \\ \hline \hline
 \multicolumn{3}{c}{\small \textbf{Singer-Songwriter}} & \multicolumn{3}{c}{\small \textbf{Screenwriter}}  \\  \hline
 \multicolumn{1}{l}{\small $n$} & \multicolumn{1}{l}{\small \textbf{$x$} (compositional)} & \multicolumn{1}{l|}{\small \textbf{$\hat{h}$} (optimized)} & \multicolumn{1}{l}{\small $n$} & \multicolumn{1}{l}{\small \textbf{$x$} (compositional)} & \multicolumn{1}{l}{\small \textbf{$\hat{h}$} (optimized)}\\ \hline 
  {\scriptsize 1 } & { \scriptsize \textbf{Eric Clapton} } & { \scriptsize \textbf{Bonnie Raitt}} &  {\scriptsize 1 } & { \scriptsize \textbf{John Lennon} } & { \scriptsize \textbf{John Lennon}} \\ 
 {\scriptsize 2 } & { \scriptsize \textbf{Bonnie Raitt} } & { \scriptsize \textbf{Eric Clapton}} &  {\scriptsize 2 } & { \scriptsize Jimi Hendrix } & { \scriptsize \textbf{Barbara Streisand}}\\ 
 {\scriptsize 3 } & { \scriptsize \textbf{Van Morrison} } & { \scriptsize \textbf{Van Morrison}} &  {\scriptsize 3 } & { \scriptsize \textbf{Barbara Streisand} } & { \scriptsize \textbf{Eric Idle}}\\ 
 {\scriptsize 4 } & { \scriptsize \textbf{B.B. King} } & { \scriptsize \textbf{B.B. King} } &  {\scriptsize 4 } & { \scriptsize Eric Clapton } & { \scriptsize \textbf{Nick Cave}}\\ 
 {\scriptsize 5 } & { \scriptsize \textbf{Bob Seger}  } & { \scriptsize \textbf{Bob Seger}} &  {\scriptsize 5 } & { \scriptsize Eddie Vedder } & { \scriptsize \textbf{Alan Bergman}}\\ \hline
 \\
 \multicolumn{3}{c}{\small \textbf{Disc Jockey} } & \multicolumn{3}{c}{\small \textbf{Writer} }  \\ \hline
 \multicolumn{1}{l}{\small $n$} & \multicolumn{1}{l}{\small \textbf{$x$} (compositional)} & \multicolumn{1}{l|}{\small \textbf{$\hat{h}$} (optimized)} & \multicolumn{1}{l}{\small $n$} & \multicolumn{1}{l}{\small \textbf{$x$} (compositional)} & \multicolumn{1}{l}{\small \textbf{$\hat{h}$} (optimized)} \\ \hline 
 {\scriptsize 1 } & { \scriptsize Tom Petty} & { \scriptsize \textbf{Steven Van Zandt} } &  {\scriptsize 1 } & { \scriptsize \textbf{John Lennon} } & { \scriptsize \textbf{Alanis Morissette} } \\
 {\scriptsize 2 } & { \scriptsize Warren Zevon } & { \scriptsize \textbf{Erykah Badu} }&  {\scriptsize 2 } & { \scriptsize \textbf{Alanis Morissette} } & { \scriptsize \textbf{John Lennon}}\\ 
 {\scriptsize 3 } & { \scriptsize Willie Nelson } & { \scriptsize \textbf{Alice Cooper} } &  {\scriptsize 3 } & { \scriptsize Paul McCartney } & { \scriptsize \textbf{Leonard Cohen} }\\ 
 {\scriptsize 4 } & { \scriptsize John Mayer } & { \scriptsize John Mayer} &  {\scriptsize 4 } & { \scriptsize Tina Turner } & { \scriptsize \textbf{Leonard Bernstein}}\\ 
 {\scriptsize 5 } & { \scriptsize Steve Earle } & { \scriptsize \textbf{Moby} } &  {\scriptsize 5 } & { \scriptsize Dolly Parton } & { \scriptsize \textbf{Prince} }\\ \hline  
 \end{tabular} 
 \caption{Semantic neighborhoods of type (pre-) and token (post-optimization) triplets output by the best-performing \textsc{HHolE} model ($d = 512, \lambda = 1.0$). \emph{US Presidents}: Effect of optimization on the semantic neighborhoods of entity embeddings in the context of the query $(\ \cdot\ , \mathtt{office\_title}, \mathtt{US\_{}President})$. \emph{Guises of Bob Dylan}: Effect of optimization on the semantic neighborhood of \texttt{Bob\_{}Dylan} in the context of four queries about his \texttt{profession}: Bob Dylan as \texttt{singer-songwriter}, \texttt{screenwriter}, \texttt{disc\_{}jockey}, and \texttt{writer}. \texttt{Bob\_{}Dylan} is a positive instance of each of these professions in FB15K. For each entity, we retrieved the 5 closest (Euclidian Distance) compositional triplet embeddings, as well as the five closest triplets, among all candidate triplets, when all these candidates are optimized.  
Triplet completions that in fact occur in FB15K are marked in \textbf{bold}. Human-readable entity names were retrieved from a mapping between Freebase machine IDs and names of Wikipedia articles built by \citet{ling2012ner}. See main text for discussion of the results.
} \label{table_sem_neighborhoods}
 \end{table*}

\section{Conclusion} \label{sec_conclusion}

In this paper, we proposed Gradient Graphs, a general method for augmenting compositional representations of Knowledge Graphs with a post-composition procedure that optimizes the well-formedness of triplet embeddings, highlighting the model's connection to Harmonic Grammar and Optimality Theory. The resulting model shows marked improvements over the compositional models it is implemented alongside, and also produces triplet token embeddings with properties that prove useful for inference about knowledge base entities. In future work, we intend to explore the utility of semantically-optimized token embeddings in other linguistic settings.

\bibliographystyle{apalike}
\bibliography{./bib/research}

\section{Appendix A: Model details }

\ex. \textbf{Claim}: $\mu(x) = -(\mathbb{W} - \lambda I)^{-1}(\frac{1}{2}b + \lambda x) $ is the unique global optimum for $\mathcal{H}(h,x)$ for any fixed $x$. \label{claim_mean_optimum}

We define the Harmony of hidden state $h$ with respect to triplet embedding $x$ as in \ref{def_harmony}:
\begin{align*}
\mathcal{H}(h,x) &\equiv \frac{1}{2} \left[  h^\top \mathbb{W} h + b^\top h   - \lambda (h - x)^\top (h-x)  \right] \\
	&= \frac{1}{2}\left[ h^\top \left(\mathbb{W} - \lambda I \right) h + (b + 2\lambda x)^\top h - \lambda x^\top x \right] \\
	&\equiv \frac{1}{2}\left[ h^\top  V h + m(x)^\top h - \lambda x^\top x \right]
\end{align*}
Completing the square yields:
\begin{flushleft}{\scriptsize
\begin{align*}
\mathcal{H}(h,x) =& \frac{1}{2}\left[ \left(h - \frac{-1}{2} V^{-1} m(x)\right)^\top V \left(h - \frac{-1}{2} V^{-1} m(x)\right) \right] \\ & + \frac{1}{2} \left[ - \lambda x^\top x - \frac{1}{4} m(x)^\top V^{-1} m(x) \right] \\
	\equiv & \frac{1}{2}\left[ \left(h - \mu(x) \right)^\top  V \left(h - \mu(x) \right)\right] + \ell(x) 
\end{align*}}\end{flushleft}
which is valid because $V = \mathbb{W}-\lambda I$ is symmetric. $\ell(x)$ does not depend on $h$, so it is sufficient to optimize $\frac{1}{2}\left[ \left(h - \mu(x) \right)^\top  V \left(h - \mu(x) \right)\right]$. Setting $\frac{\partial \mathcal{H}(h,x)}{\partial h} = 0$ yields $2V(h - \mu(x)) = 0$; $\therefore h = \mu(x)$. Since $V$ is negative-definite, this point is a \emph{maximum}. 

The truth of the claim may be more quickly perceived by observing that $\mathcal{H}(h,x)$ defines a Gaussian distribution over the hidden state variable $h$ with mean $\mu(x)$ and precision matrix $\Sigma^{-1} \equiv -V$. The optimality of $\mu(x)$ then follows from the unimodality of Gaussians. 

The training objective \ref{def_harmony_loss} may be justified by the following considerations. We take the compositional triplet data to be generated by hidden states of the gradient graph network, and maximize the log probability of the training data using the maximum a posteriori point estimate of the hidden state $h$. The "complete data" are then $\mathcal{D} = \{ \langle \hat{h}, x\rangle \} = \{ \langle \mu(x), x \rangle \}$. For fixed $x$, $\mathcal{H}(h,x)$ models the conditional distribution $p(h|x)$, with 
\ex. $p(h|x) = \frac{\exp\{\mathcal{H}(h,x)\}}{Z(x)}$

where $Z(x) = \int_{h'} \exp \left\{\mathcal{H}(h',x)\right\} dh' = |2\pi V^{-1}|^{\frac{1}{2}}\exp\{\ell(x)\} $ is the partition function conditioned on $x$.
Let $\chi_q = \{x'\}$ be the set of candidate triplet embeddings consistent with a given query $q$. Choosing the discrete distribution $p(x) = \frac{\exp\{\ell(x)\} }{\sum_{x'\in \chi_q} \exp \{ \ell(x')\}}$ over triplet embeddings as the prior probability of the embedding $x$,\footnote{Note that, when $\mathcal{H}(\cdot,x)$ is evaluated at $\mu(x)$, the only nonzero term in $\mathcal{H}(\mu(x),x)$ is $\ell(x)$. Hence, it is sufficient to perform gradient descent on the prior: $\ell(x)$}
 we have:
\begin{align*}
p(\mu(x),x|q) \propto& \frac{\exp \{ \mathcal{H}(\mu(x),x)\} }{|2\pi V^{-1}|^{\frac{1}{2}}\sum_{x'\in \chi_q} \exp \{ \ell(x')\}} 
\end{align*}
For given parameters, the denominator is constant. So, renormalizing over the discrete triplets $\chi_q$ gives:
\begin{align*}
p(\mu(x),x|q) = \frac{\exp \{ \mathcal{H}(\mu(x),x)\} }{\sum_{x'\in \chi_q} \exp\{\mathcal{H}(\mu(x),x)\}}
\end{align*}
Approximating the discrete distribution over all of $\chi_q$ with a negative sample yields the objective \ref{def_harmony_loss}. 

\section{Appendix B: Implementation details}

In initial experiments, we searched through a number of candidate models. These included two Harmonic variants of the \textsc{Rescal} model \citep{nickel2011rescal}, as well as models that constructed $x$ as a simple concatenation of entity and relation vectors, as well as three-way tensor products of these vectors. These initial experiments led us to focus on \textsc{DistMult} and \textsc{HolE} as the best-performing candidates. 
Our Harmonic models and reimplementations of the \textsc{DistMult} and \textsc{HolE} baselines were written in TensorFlow \citep{tensorflow} and estimated using the Adam optimizer \citep{adamoptimizer}. With the exception of the \textsc{HolE} reimplementation, we uniformly used the log-softmax loss \ref{def_harmony_loss}, which performed best in initial experiments. In contrast, \citet{yang2015distmult} use a margin-based ranking loss that is linear in the margin between the scores of positive and negative examples up to a threshold, and \citet{nickel2016hole} use the pairwise linear margin loss applied to the scores squashed by the logistic function. For \textsc{HolE}, we used the \emph{linear} margin loss, which provided by far the best performance in the experiments. For each model, we trained until performance on the validation set decrease, then chose the best-performing embedding size from among $d\in \{256,512\}$. Batch size (512), negative sampling rate (500), and learning rate (0.001) were kept constant across models. We note in passing that regions of the hyperparameter space for \textsc{DistMult} explored by \citet{kaldec2017baselines} were inaccessible to us for technical reasons. For the Harmonic models, we manually tuned the $\lambda$ hyperparameter.

\end{document}